# Generalizing the Kelly Strategy

Arjun Viswanathan, Head, Rates Big Data,
Citibank Global Markets Limited (CGML)

v4. Dec 5, 2016

**Abstract**

A recent draft by Victor Haghani and Richard Dewey [1] describes an experiment where participants were given initial wealth, a coin of known bias, and could bet a (variable) proportion of their in-game wealth on a sequence of flips of this coin. Assuming log utility and uncapped reward, the optimal strategy is to bet as per the Kelly criterion. Interestingly, the participants in general did not do so - many bet a larger proportion, up to 100% of their game assets. This note shows that such behaviour can be rational, if one takes into account the effect of extraneous wealth. The optimal solution for log utility with extraneous wealth is found, and extended to the optimal solution for a wide class of utility functions. A counterintuitive result is proved : for any continuous, concave, differentiable utility function, the optimal choice at every point depends *only on the probability of reaching that point*. That is, the optimal choice depends *only on the number of heads encountered*, regardless of the sequence. Lastly, the practical calculation of the optimal bet at every stage is made possible through use of the binomial expansion, reducing the problem size from exponential to quadratic. This makes the solution practical for games with many hundreds of steps.

The author thanks Vlad Ragulin[1], for introducing the original problem, and Andy Morton[2] for motivating investigation of the general case and economic interpretation of the results.

---

[1][Director, US Govt Bond Trading, CGML]
[2][Global Head G10 Rates, Markets Treasury and Finance, CGML]

# 1 Setup

The player is given 1 unit of game wealth, and the chance to bet on $f$ flips of a biased coin coin with known probability $p > 0.5$ of coming up heads. At every flip, the player may bet a proportion $b \in (0,1)$ of their current game wealth $g$ on heads. The player also has extraneous wealth $w$, which is not affected by the betting.

We initially assume the player's internal reward function is log utility, i.e. they aim to maximise $E[Log(g_{final} + w)]$ where $g_{final}$ is their final game wealth. If optimizing only one step ahead, the optimal bet $= (2p-1)(1 + \frac{w}{g})$, which reduces to the Kelly betting criterion $(2p-1)$ if $w = 0$.

Interestingly, this is not optimal for an n-step game. As one would expect, as $g \gg w$, the game wealth dominates and the optimal bet converges to $(2p-1)$. If $g \ll w$, *and the game is due to end soon*, the optimal bet is 1.00, which agrees with the intuition that there is little downside to betting the entire tiny stake.

But for moderately long games, the risk of losing paths that could lead to large wealth implies that the first bet must be lower than 1. For example, if p = 0.6 and w = 1000, for 25 flips the optimal first bet is approximately 0.659.

# 2 Convenient notation

The possible paths of the game form a complete binary tree. We number the nodes of this tree with the root = 1.

Node $n$ has children Node $2n$ (if the coin gives heads) and Node $2n+1$ (if the coin gives tails).
For $f$ flips, the final level (that is, level $f + 1$) has $2^f$ nodes. Let

$$g_n = \text{the game wealth at node n}$$

$$b_n = \text{the bet at node n}$$

$$p_n = \text{the probability of reaching node n} \tag{1}$$

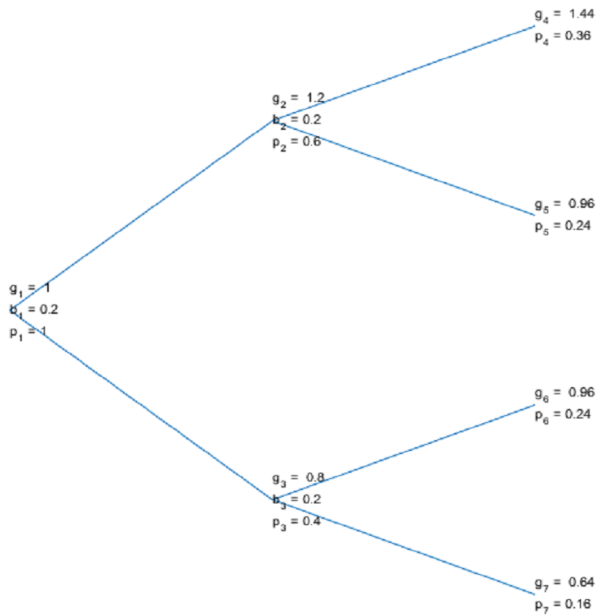

Figure 1: 2-flip game, $p = 0.6$, player bets according to the Kelly criterion

Node n belongs to level $\lfloor Log_2(n) \rfloor + 1$, that is $Floor[Log_2(n)+1]$. At level m, $p_k$ may take m+1 distinct values, corresponding to the $m+1$ possibilities $0, 1, ..., m$ of the number of heads in $m$ flips. Evidently $g_n \geq 0$ for all $n$. Also $g_{2n} = g_n(1+b_n)$ and $g_{2n+1} = g_n(1-b_n)$. This gives us the following recurrences:

$g_n = \frac{g_{2n} + g_{2n+1}}{2}$

$b_n = \frac{g_{2n} - g_{2n+1}}{g_{2n} + g_{2n+1}}$

(2)

This is useful : if we know the final game wealths, we know for free all previous bets and wealths.

## 3 Analytic optimal solution for log utility & extraneous wealth

Since $g_1 = 1$ and by (2), the unweighted sum of wealths at level j equals $2^{j-1}$

We seek to maximize the total utility at the final level

$\sum_{k=2^f}^{2^{f+1}-1} p_k Log(g_k + w)$

With additional equality constraint

$\sum_{k=2^f}^{2^{f+1}-1} g_k = 2^f$

And inequality constraint

$g_k \geq 0$ for $k = 2^f, 2^f + 1, ..., 2^{f+1} - 1$

This is an optimization problem with convex and differentiable objective, affine equality constraint, and convex inequality constraint. Therefore the KKT conditions [**2**] are necessary and sufficient to find an optimum. Let x (with $2^f$ entries) be the solution of this problem (x is the vector of final wealths)

Introducing Lagrange multipliers $\lambda$ (a vector with $2^f$ entries) for the inequality constraint, and $\nu$ (a scalar), for the equality constraint, we get the standard KKT conditions :

$x_k \geq 0, \quad \sum_{k=1}^{2^f} x_k = 2^f, \quad \lambda_k \geq 0, \quad \lambda_k x_k = 0,$

and lastly $\frac{-p_k}{x_k + w} - \lambda_k + \nu = 0$, for $k = 1, 2, ..., 2^f$

Usefully, the above can directly be solved for x. Eliminating $\lambda_k$ between the last 2 conditions gives :

$(\nu - \frac{p_k}{x_k + w}) x_k = 0$ and $\frac{p_k}{x_k + w} \leq \nu$

Now suppose $\nu < \frac{p_k}{w}$. Then $\frac{p_k}{x_k + w} \leq \nu < \frac{p_k}{w}$ which can only occur if $x_k > 0$. But then, since

$(\nu - \frac{p_k}{x_k + w}) x_k = 0$ we must have $\nu = \frac{p_k}{x_k + w}$, which implies $x_k = \frac{p_k}{\nu} - w$

Conversely suppose $\nu \geq \frac{p_k}{w}$. But then $x_k$ cannot be strictly greater than zero, for if it was,

$(\nu - \frac{p_k}{x_k + w}) x_k$ is the product of two terms, both strictly greater than 0, and thus cannot be 0.

Therefore we have $x_k = \frac{p_k}{\nu} - w$ if $\frac{p_k}{\nu} - w > 0$, and 0 otherwise, ie $x_k = max(0, \frac{p_k}{\nu} - w)$

Substituting $x_k$ into the equality constraint, we have

$$\sum_{k=1}^{2^f} max(0, \frac{p_k}{\nu} - w) = 2^f \tag{3}$$

The LHS of which is a strictly decreasing continuous function of $\nu$, is zero for large enough $\nu$, and also arbitrarily large if $\nu$ is small enough. By the Intermediate Value Theorem the equation has a unique solution (which can be readily determined, e.g. by binary search).

Once we have $\nu$, we have $x$, the vector of optimal final wealths. We then use (2) to get the optimal bets at every node. Interestingly, we see that the value of $x_k$ depends only on $p_k$ or equivalently the number of heads in f flips.

This holds in the general case, as is proved shortly, which provides a route to efficient calculation of the best bet at any point.

## 4 Example

Following the strategy above for a 4-flip game with $p = 0.6$, $w = 20$ would yield the following game tree. Paths where the player bets all their game wealth and loses are greyed out.

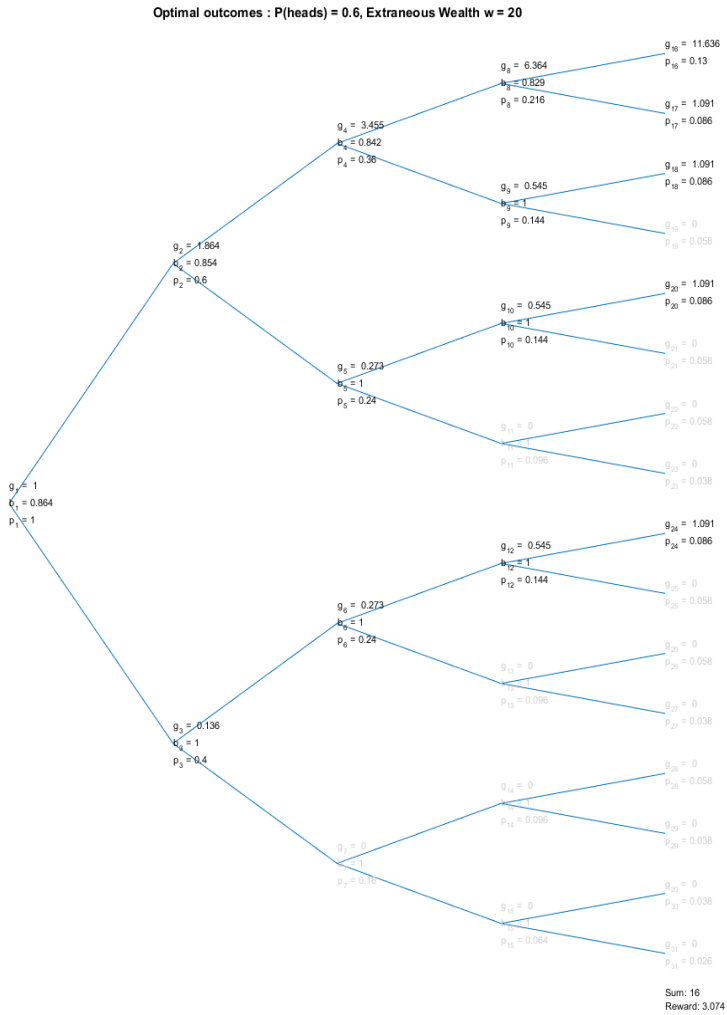

Figure 2: 4-flip game, $p = 0.6$, w = 20, optimal bets diverge from the Kelly strategy

# 5 Analytic solution for more general utility functions

We now consider a general utility function $F : \mathbb{R} \to \mathbb{R}$, reasonably be assumed to be continuous, concave and differentiable. Linear and log utility are special cases of this type. The derivative $F'$ need not be continuous, although it is continuous almost everywhere. $F$ is concave, therefore $F'$ is monotonically decreasing. We therefore know :

$F'(w) \geq F'(x_k + w)$. $F'$ attains a (not necessarily unique) max at $w$ and a (not neccessarily unique) min at $w + 2^f$, and if the strict inequality $F'(w) > F'(x_k + w)$ holds then we must have $x_k > 0$

We now follow exactly the same proof pattern as before, with $F(x_k + w)$ and $F'(x_k + w)$ in place of $Log(x_k + w)$ and $\frac{1}{x_k)+w}$

As before we seek to maximize

$\sum_{k=2^f}^{2^{f+1}-1} p_k F(g_k + w)$

subject to the constraints

$\sum_{k=2^f}^{2^{f+1}-1} g_k = 2^f \quad \text{and} \quad g_k \geq 0 \quad \text{for } k = 2^f, 2^f + 1, ..., 2^{f+1} - 1$

As before, introducing Lagrange multipliers $\lambda$ (a vector with $2^f$ entries) for the inequality constraint, and $\nu$ ( a scalar) for the equality constraint, we get the standard KKT conditions :

$x_k \geq 0, \quad \sum_{k=1}^{2^f} x_k = 2^f, \quad \lambda_k \geq 0, \quad \lambda_k x_k = 0,$

and lastly $-p_k F'(x_k + w) - \lambda_k + \nu = 0$ , for $k = 1, 2, ..., 2^f$

Eliminating $\lambda_k$ between the last 2 conditions gives :

$(-p_k F'(x_k + w) + \nu) x_k = 0$ and $p_k F'(x_k + w) \leq \nu$

The first term is the product of two nonzero terms. If either is strictly positive, the other is zero. As before we will use this to determine $x_k$. Define the function H as follows:

$H(y) =$ the largest total wealth $z \in [-\infty, w + 2^f]$ for which $F'(z) \geq y$

H is decreasing and is roughly the inverse function of the derivative of the utility: given $y$, $H[y]$ gives a total wealth $\leq w + 2^f$ where the marginal utility is as close to $y$ as can be without dropping below $y$.

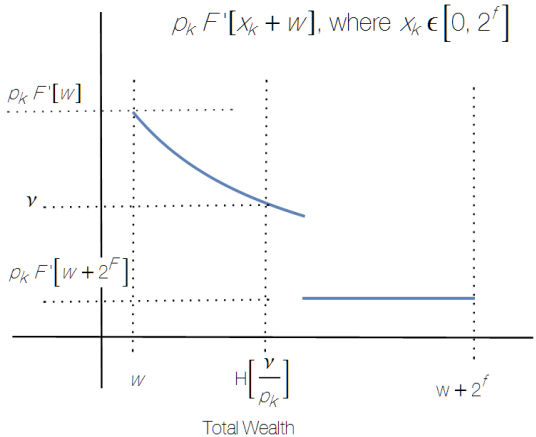

Figure 3: illustration of F' and H, where F is capped log utility.

Now suppose $\nu < p_k F'(w)$

Then $p_k F'(x_k + w) \leq \nu < p_k F'(w)$, which can only occur if $x_k > 0$

In this case we can solve for $\nu$ and $x_k$ :

$$-p_k F'(x_k + w) + \nu = 0 \quad \Rightarrow \nu = p_k F'(x_k + w) \quad \Rightarrow x_k = H(\tfrac{\nu}{p_k}) - w$$

Secondly suppose $\nu > p_k F'(x_k + w)$. Then $H(\tfrac{\nu}{p_k}) - w$ must be $< 0$

For if not, then there exists some $z \geq w$ for which $p_k F'(z) \geq \nu > p_k F'(w)$, which cannot be as $F'$ is decreasing.
So $H(\tfrac{\nu}{p_k}) - w < 0$ and consequently $-p_k F'(w) < \nu$. But then

$$\nu - p_k F'(x_k + w) \geq \nu - p_k F'(w) > 0 \quad \Rightarrow x_k = 0$$

Finally suppose $p_k$F'(w) $= \nu$, in which case $x_k$ might be zero or positive. Again we set $x_k = H(\tfrac{\nu}{p_k}) - w$
= the largest $x_k$ for which $F'(x_k + w) = F'(w)$ (equality holds as $F'$ is decreasing).
In either case, $x_k$ satisfies the KKT conditions.

Therefore we have

$$x_k = H(\tfrac{\nu}{p_k}) - w \text{ if } H(\tfrac{\nu}{p_k}) - w > 0 \text{ and } 0 \text{ otherwise, i,e } x_k = max(0, H(\tfrac{\nu}{p_k}) - w) \tag{4}$$

Substituting $x_k$ into the equality constraint, we have

$$\sum_{k=1}^{2^f} max(0, H(\tfrac{\nu}{p_k}) - w) = 2^f \tag{5}$$

The LHS of which is a decreasing function of $\nu$, and can be efficiently solved for $\nu$ , e.g. by binary search). We note the symmetry between the above expressions and the expressions for the optimal solution for log utility.

Once we have $\nu$, we have $x$, the vector of optimal final wealths. We then use (2) to get the optimal bets at every node.

## 6 Efficient Calculation

There are only f+1 distinct values of $p_k$, namely $p^f, p^{f-1}(1-p), ..., (1-p)^f$
We can use the binomial expansion [3] of equation (5)

$$\sum_{k=1}^{2^f} \max(0, H(\tfrac{\nu}{p_k}) - w) = \sum_{j=0}^{f} \binom{f}{j} max(0, H(\tfrac{\nu}{p^j(1-p)^{f-j}}) - w) = 2^f$$

to calculate $\nu$. This conveniently reduces our effort from exponential to linear to calculate $x$.
Moreover, we can recast our complete binary tree with $2^m$ nodes at level $m$, into a recombining tree with $m$ nodes at level $m$. Specifically, if $node_{mn}$ is the $n$th node at level $m$ of the recombining tree, let $g_{mn}$, $b_{mn}$, $p_{mn}$, be the game wealth at, bet at, and probability of arriving at, $node_{mn}$ .
?

Then, by (2) , we have the recurrences

$$g_{mn} = \frac{g_{(m+1)n} + g_{(m+1)(n+1)}}{2}$$

$$b_{mn} = \frac{g_{(m+1)n} - g_{(m+1)(n+1)}}{g_{(m+1)n} + g_{(m+1)(n+1)}} \tag{6}$$

Each level of $m$ nodes takes $O(m)$ effort to calculate. The full set takes effort $O(1+2+...+f) = O(f^2)$
This reduces the effort for the full tree from exponential to quadratic:

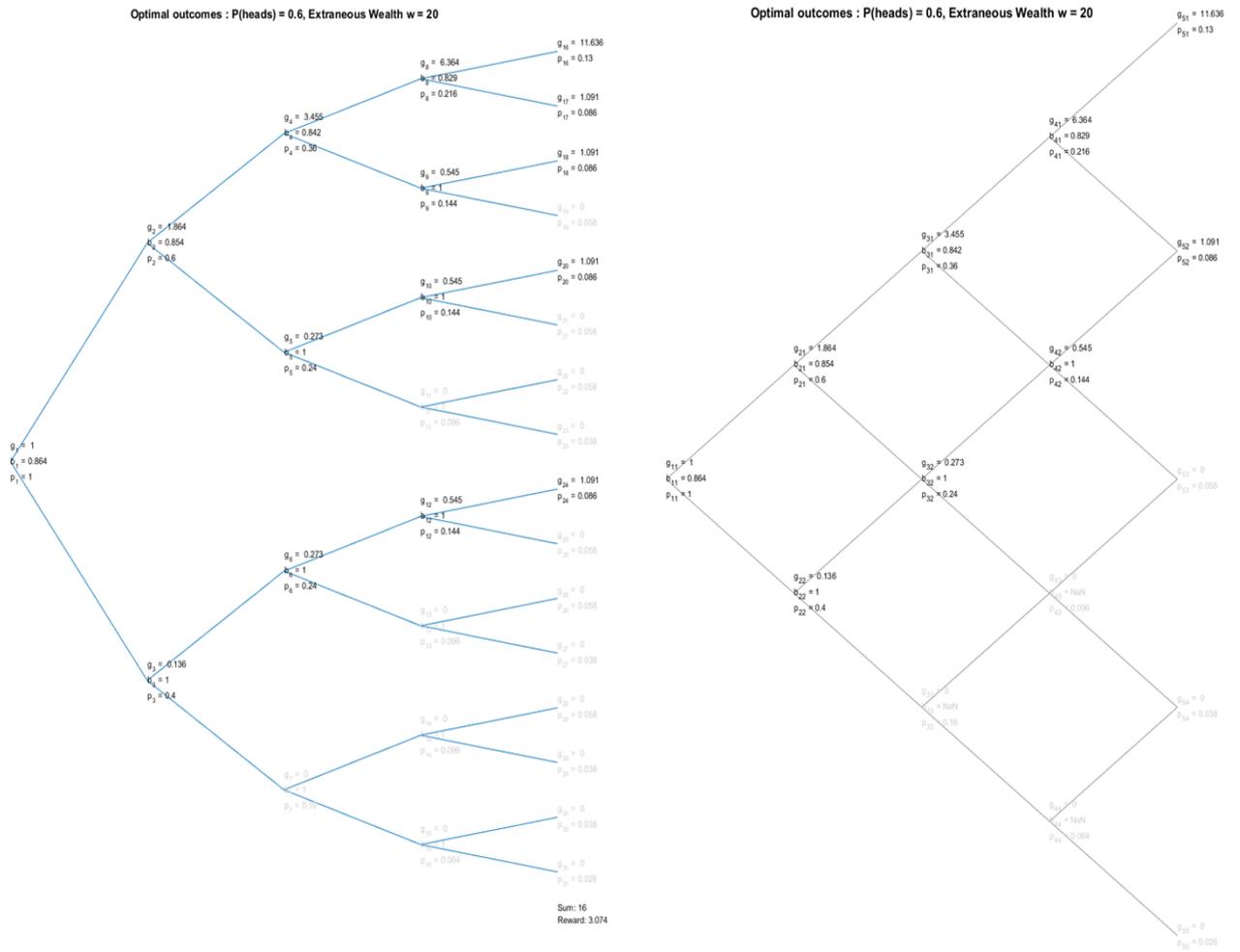

Figure 4: Binary and recombining trees

The expression for the very first bet has a nice form : if $x_0, x_1, ...x_f$ are the *distinct* values that the final wealth can take in the *recombining* tree, the very first bet is

$$\frac{1}{2^f} \sum_{j=0}^{f} \binom{f}{j} \frac{f-2j}{f-j} x_j$$

Which may be verified in a variety of ways: induction on f, binomial identities, generating functions. [3]

# 7 Conclusion

This seems a natural point to pause. This note has generalized the standard Kelly strategy to an optimal strategy for situations with extraneous wealth, and under a much more general class of reward functions.

The optimal strategy for pathological reward functions e.g. nowhere differentiable, are left as an exercise to the reader, but are unlikely to occur in reality (one hopes).

Of practical value is the calculation methodology which lets us efficiently derive optimal strategies for games of many hundreds or thousands of turns. The biased coin metaphor applies not only to long term investing but any decision process where entities have a notional edge- for example, to help a market making desk create prices more optimally.

This method might prove of use in more complex games, e.g. the Halite Artificial Intelligence Challenge at www.halite.io , which shares some key similarities with the coinflipping game.

The note concludes with an anecdote:

*A mathematician was asked by their manager to design a chair. They quickly solved the problem of a chair with zero legs, and with slightly more effort, a chair with infinite legs. A chair with positive real (though not necessarily integer) legs was conjectured to exist. Finally, one weekend, they solved this variation as well, although negative legs remains an open problem.*